\documentclass[letterpaper,10pt,conference]{ieeeconf}  

\IEEEoverridecommandlockouts                              
\usepackage{graphicx} 
\usepackage{amsmath} 
\usepackage{amssymb}  
\usepackage{algorithm}
\usepackage{algpseudocode}
\usepackage{xcolor}

\makeatletter
\let\NAT@parse\undefined
\makeatother
\usepackage{hyperref}
\usepackage{float}
\graphicspath{{figures/}}
\usepackage{graphicx}
\usepackage{subcaption}
\usepackage{wrapfig}

\title{\LARGE \bf From Cooking Recipes to Robot Task Trees -- Improving Planning Correctness and Task Efficiency by Leveraging LLMs with a Knowledge Network}

\author{Md. Sadman Sakib, and Yu Sun\\
\thanks{
Yu Sun and Md. Sadman Sakib are members of the Robot Perception and Action Lab (RPAL), which is part of the Department of Computer Science \& Engineering at the University of South Florida, Tampa, FL, USA. 
\newline\textit{Email}: \texttt{\{mdsadman,yusun\}@usf.edu}}%
}

\begin{document}

\maketitle

\thispagestyle{empty}
\pagestyle{empty}

\begin{abstract}

Task planning for robotic cooking involves generating a sequence of actions for a robot to prepare a meal successfully. This paper introduces a novel task tree generation pipeline producing correct planning and efficient execution for cooking tasks. Our method first uses a large language model (LLM) to retrieve recipe instructions and then utilizes a fine-tuned GPT-3 to convert them into a task tree, capturing sequential and parallel dependencies among subtasks. The pipeline then mitigates the uncertainty and unreliable features of LLM outputs using task tree retrieval. We combine multiple LLM task tree outputs into a graph and perform a task tree retrieval to avoid questionable nodes and high-cost nodes to improve planning correctness and improve execution efficiency. Our evaluation results show its superior performance compared to previous works in task planning accuracy and efficiency.

\end{abstract}

\section{Introduction}

Robotic cooking has emerged as a highly promising domain within robotics, presenting notable advantages such as convenience and the potential for enhanced efficiency and precision in meal preparation. To effectively automate cooking tasks, the key component is efficient task planning. This entails generating a series of actions guiding the robot in accomplishing a specific goal. However, this is an intricate field of research due to the fact that cooking tasks typically involve lengthy sequences of actions encompassing various ingredients and tools. Moreover, they necessitate the attainment of numerous crucial ingredient states throughout the process. Additionally, the cooking conditions, processes, and requirements are exceptionally diverse. Approaches like state-space planning, learning from demonstration, and even knowledge network retrieval encounter difficulties when confronted with unseen starting conditions and requests.

In cooking tasks, ingredients or objects can vary in form, shape, and size, and there are multiple states to consider during recipe execution. The manipulation of an ingredient depends on its specific state, and certain ingredients may not be readily available in the desired state. Additionally, robots have varying capabilities, making some actions easier for them to perform than others. 
A task planning method should consider these factors and propose a plan that is most suitable for the robot to execute efficiently. Previous work has created a knowledge network consisting of 140 cooking recipes called the Functional Object-Oriented Network (FOON) \cite{paulius2016functional, sakib2021evaluating}. 
However, generating plans in novel scenarios where FOON lacks knowledge about the recipe or an ingredient proved challenging. Furthermore, expanding the knowledge base was difficult due to the reliance on manual annotation. 

Recently, the emergence of Large Language Models (LLMs) \cite{Brown2020LanguageMA, devlin2018bert, Wei2021FinetunedLM} has addressed the limitation of limited knowledge. These LLMs possess the ability to generate ``likely'' viable solutions for different scenarios and requests. While their results may not always be correct or optimal, their notable capacity for generalization can help overcome the limitations of search-based task tree retrieval methods. The search-based retrieval approach with a comprehensive knowledge network on the other hand can detect, remove and replace the wrong elements in the LLM outputs. 

\begin{figure}[!ht]
	\centering
    \includegraphics[width=\columnwidth]{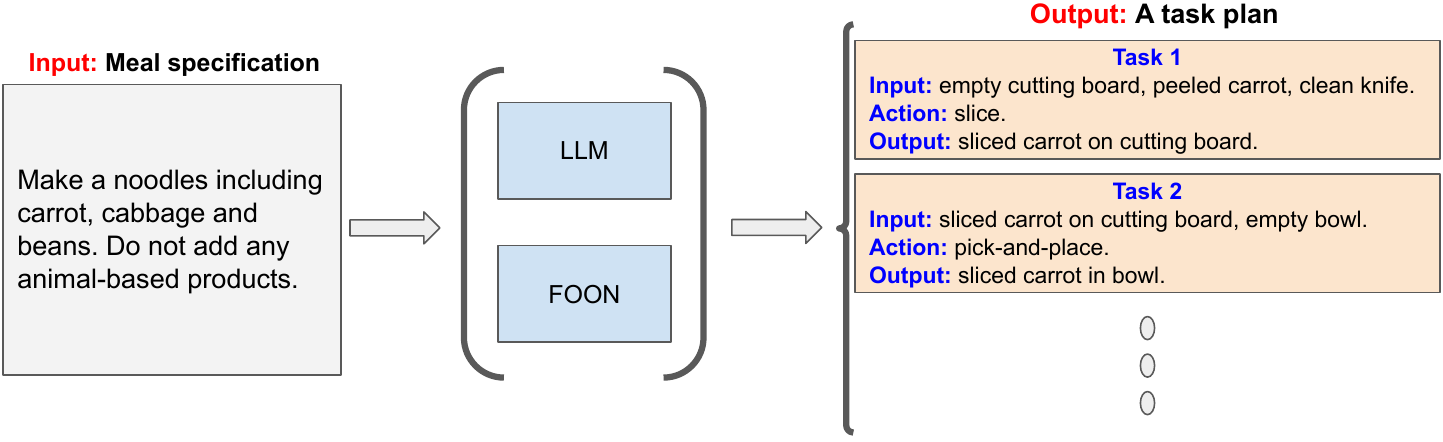}
	\caption{
        Overview of our approach. Given a meal preparation instruction, the model generates a list of tasks specifically designed for the robot. 
	}
	\label{fig:intro}
\end{figure}

The primary focus of this research paper is to tackle the task planning challenge in robotic cooking through the introduction of an innovative task tree generation approach (Figure \ref{fig:intro}). We aim to generate a task plan that is both error-free and cost-effective.
To enhance the accuracy of the task plan, we employ a method that involves detecting incorrect components within the task trees generated by GPT-3 and search for alternative options either within other task trees or within the FOON knowledge graph. This approach allows us to improve the overall quality and reliability of the generated task plan.
By carefully selecting the most optimized plan from these alternatives, the pipeline ensures effective resource utilization while achieving the desired objectives. The effectiveness of the task tree generation pipeline is evaluated through a comparative analysis with a previous approach. The results demonstrate the superiority of the proposed method, showcasing enhanced task planning accuracy and improved cost-efficiency. 
%

Our contributions in this paper are as follows: (i) We propose a novel task tree generation approach that accepts any dish of the user's choice and produces a robot task tree with state-of-the-art accuracy and efficiency; (ii) We fine-tune GPT-3 to convert natural language instructions into a task tree structure; (iii) We improve the accuracy of the task plan by detecting incorrect components in the GPT generated task trees and finding alternatives either in other task trees or FOON; (iv) We optimize execution costs by performing weighted retrieval in a mini-FOON combined from multiple GPT outputs or FOON. We demonstrate the superiority of our model through a comparison with a previous approach.

\section{Background}

\subsection{Functional Object-Oriented Network}

\label{sec:FOON}
\begin{figure}[!ht]
	\centering
    \includegraphics[width=\columnwidth]{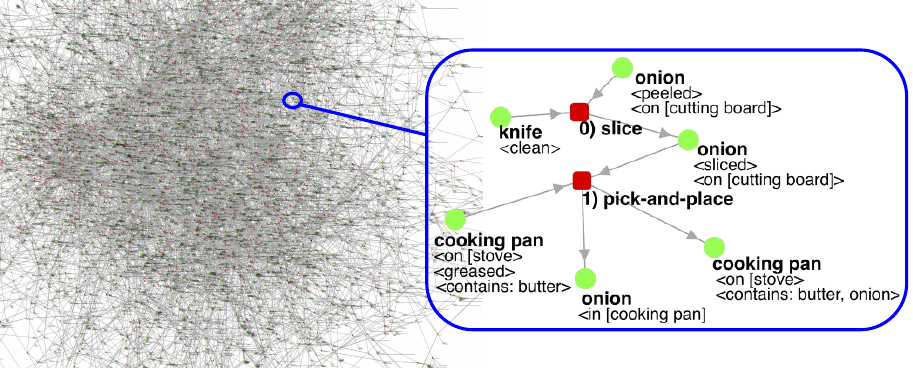}
	\caption{
	Two functional units from FOON depicting slicing an onion and placing it to the cooking pan. Objects and motions nodes are denoted by green circle and red square respectively. 
	}
	\label{fig:unit}
\end{figure}

FOON and related knowledge graphs have been used in many tasks for robots, such as robotic cooking \cite{dualarm} and furniture assembly \cite{ikeabot, ikeaasm, onto_ikea}. The one used here is a knowledge graph constructed through manual annotation of video demonstrations. It consists of two types of nodes: object nodes and motion nodes. These nodes are connected by directed edges, which depict the preconditions and effects of actions.
The functional unit is the fundamental building block of FOON, representing a single action observed in the video demonstration. It consists of one or more input nodes, one or more output nodes, and a single motion node. The input nodes specify the required state of objects before the action, while the output nodes describe the resulting state after the action is executed. The motion node represents the action itself. Functional units provide a detailed and vivid representation of the actions observed in the video demonstrations.
Figure \ref{fig:unit} shows two functional units of slicing an onion and placing onion to cooking pan. 
The current FOON dataset (available in ~\cite{foonet}) consists of 140 annotated recipes sourced from platforms such as YouTube, Activity-Net \cite{caba2015activitynet}, and EPIC-KITCHENS \cite{Damen2018EPICKITCHENS}.

\subsubsection{Task Planning with FOON}

The utilization of FOON as a knowledge base for task planning offers several advantages, including the ability to provide recipe variations. Task planning with FOON involves searching the network to find a goal node and retrieving a path, referred as a task tree, that leads to achieving the desired objective. A task tree, consists of a sequence of functional units that need to be executed in order to prepare the dish. To illustrate, consider the task tree associated with boiling water, which comprises actions such as placing a pot on the stove,
pouring water, turning on the stove, and turning off the stove. Each of these procedural steps is represented by input object nodes, signifying the prerequisites for executing the action; a motion node, denoting the action itself; and output object nodes, denoting the effect of executing the action.
The task tree retrieval algorithm proposed in \cite{paulius2016functional} focuses on finding a path that utilizes only the ingredients available in the kitchen. On the other hand, \cite{paulius2021weighted} retrieves a plan that can be executed with human-robot collaboration. 
Nevertheless, these approaches have a limitation when it comes to generating a plan for a recipe that is not explicitly available in FOON. For example, if a user asks for a plan to prepare a mango milkshake, but there is no dedicated recipe for it in FOON, the system may be unable to provide a plan, even if there is a recipe for a banana milkshake.
To address this limitation, a novel task tree retrieval method \cite{Sakib2022ApproximateTT} was introduced that can learn from similar recipes in FOON and make necessary modifications to match the user's requirements. While this approach introduces some level of generalization, the quality of the generated plan heavily relies on the availability of closely matched recipes in FOON. In this work, we leverage LLMs to overcome this dependency on closely matched recipes and generate high-quality task trees for any recipe, thereby enhancing the flexibility and effectiveness of the task planning process.

\subsection{Related Works}

In the domain of robotic cooking and task planning, several strategies have been proposed to tackle the challenges associated with generating effective action sequences for executing user instructions. One prominent approach involves the use of knowledge graphs to address this challenge. Notably, the KNOWROB framework \cite{Beetz2018KnowR2, pancake} has made significant contributions in this area by leveraging a knowledge base constructed from data collected in sensor-equipped environments. \cite{Daruna2021TowardsRO} introduced a task generalization scheme that relaxes the requirement of having multiple task demonstrations to perform tasks in unknown environments. This scheme integrates the task plan with a knowledge graph derived from observations in a virtual simulator. The impact of knowledge graphs on a robot's decision-making process was further investigated in \cite{Daruna2022ExplainableKG}. However, these approaches heavily rely on the limited information contained in their respective knowledge bases. In contrast, our approach harnesses the power of Language Models (LLMs) to alleviate the burden of creating a knowledge base, offering a more comprehensive and flexible solution.

Recently, task planning with LLMs has become a prominent area of research, capitalizing on the impressive language understanding and generation capabilities of LLMs. Various studies have explored the use of LLMs to generate step-by-step plans for long-horizon tasks. For instance, Erra et al. \cite{erra} proposed an approach that employs LLMs to generate plans for complex tasks. \cite{Huang2022LanguageMA, Shah2022LMNavRN, Zeng2022SocraticMC} have also utilized LLMs for plan generation in different domains.
However, these works often do not explicitly consider the robot's capability to perform specific actions. One limitation of relying solely on LLMs is the lack of interaction with the environment. To address this limitation, SayCan \cite{saycan} introduced a framework that combines the high-level knowledge of LLMs with low-level skills, enabling the execution of plans in the real world. By grounding LLM-generated plans with the robot's capabilities and environmental constraints, SayCan bridges the gap between language-based planning and physical execution.
In addition, recent research efforts such as Text2Motion \cite{lin2023text2motion}, ProgPrompt \cite{singh:progprompt} have integrated LLMs with learned skill policies. 
They exhibit trust in the LLM-generated plan and proceed to execute it whereas we focus on enhancing LLM's accuracy to generate an optimal task plan.




\section{Proposed Method}

\begin{figure}[ht]
	\centering
	\includegraphics[width=\columnwidth]{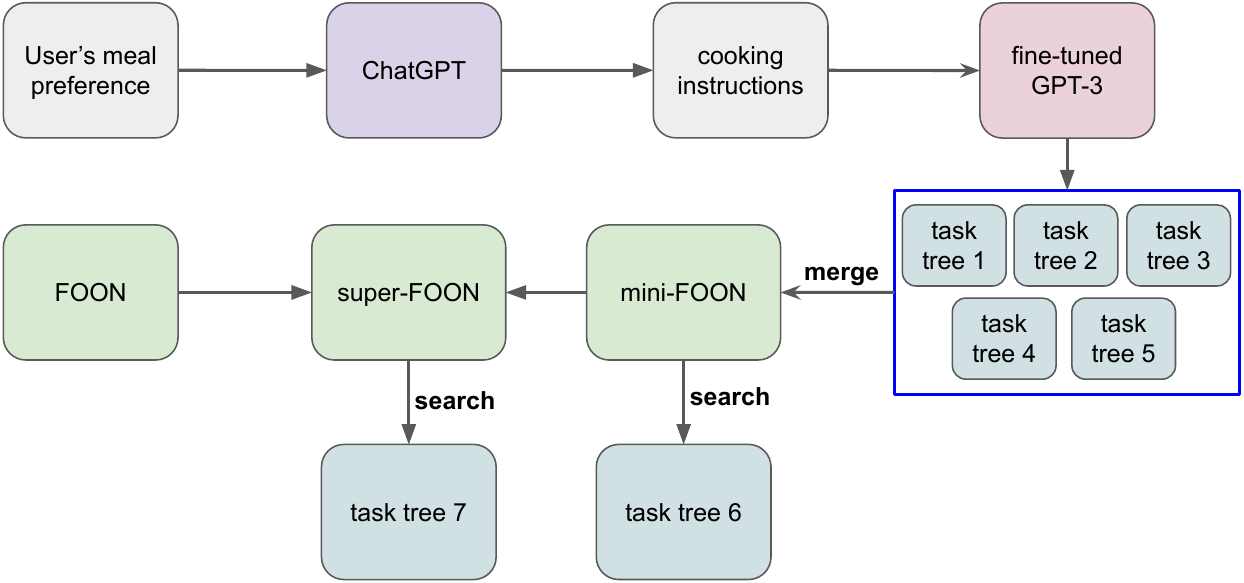}
	\caption{Overview of our task tree generation procedure. Starting with a meal specification as the query, our pipeline generates a task plan represented as task tree 7.}
	\label{fig:pipeline}
\end{figure}

Our objective is to develop a robust pipeline that generates highly accurate and executable task trees for robotic operations. To achieve this, we employ a multi-step approach that leverages the capabilities of LLMs and FOON.
Initially, we utilize ChatGPT~\cite{chatgpt} to generate a recipe based on the user's meal specifications. However, the output is in natural language, which may pose challenges for direct robot execution. To address this, we employ a fine-tuned GPT-3 model to convert the recipe instructions into a task tree format. Due to the uncertainty of the generative model, the task plan may not be always correct or most efficient.  
To enhance reliability and efficiency, we look for alternative options in other task trees generated by GPT-3 or in FOON. 
From these alternatives, the selected task tree is expected to be accurate and easier for the robot to execute.
%
%
%
A visual representation of our pipeline and its key components are presented in  Figure~\ref{fig:pipeline}. In the following subsections, we will provide detailed explanations of each component.

\subsection{Prompt engineering for recipe generation}

\label{subsection1}
Our system is designed to accommodate dish specifications provided by the user. The user can specify a list of desired ingredients or exclude certain ingredients. Additionally, specifications such as gluten-free, vegetable-based, or non-dairy options are also accepted. Based on this information, we engine a prompt and retrieve the recipe from ChatGPT. 
To facilitate easier parsing, we have designed the prompt to include numbered instructions within the response from ChatGPT. 



\subsection{Converting instructions to a task tree}

When a robot performs an action, several factors need to be considered, such as preconditions, effects, and the objects involved. Additionally, understanding the state of these objects is crucial for determining the appropriate grasp or manipulation technique. However, extracting all this information from a recipe written in natural language poses significant challenges. To address this complexity, we propose translating the instructions into structured functional units that encapsulate all the necessary details.
By organizing these functional units into a task tree, we provide a step-by-step guide for the robot to execute the task effectively. To accomplish this, we have created a dedicated dataset for fine-tuning a GPT-3 Davinci model. This model takes a recipe as input and translates it into a task tree representation. The dataset comprises 180 recipe examples sourced from FOON, each consisting of natural language instructions and a corresponding FOON task tree.
Due to the limitation of maximum token count, some recipes had to be divided into multiple parts, resulting in multiple task trees for a single recipe. 

\subsection{Creating a mini-FOON}

To address the potential presence of errors in the task plans generated by the fine-tuned model, we adopt a strategy of generating multiple task trees for the same recipe. Our aim is to search for a task tree that is error-free and one that is efficient for the robot to execute from the combined graph mini-FOON. 
FOON has revealed that merging recipes in a graph structure can lead to the emergence of novel cooking methods. This merging process allows recipes to share information and learn diverse approaches for accomplishing subtasks. Inspired by this idea of exploring new paths, we employ a similar graph structure to merge the five task trees generated by GPT. This merged structure is referred to as a mini-FOON. 

\subsubsection{Merging task trees}

During the merging process, our objective is to eliminate any incorrect functional units and remove duplicates. An incorrect functional unit can arise in two ways: (i) syntax error and (ii) an erroneous object-action relationship. Syntax verification involves checking whether the functional unit includes the necessary components such as input and output objects, as well as a motion node. Additionally, it verifies if each object has an assigned state. On the other hand, validating the object-action relationship poses the challenge of determining if the state transition for an action is correct. To tackle this challenge, we compiled a comprehensive list of all valid state transitions from FOON. Based on this list, we can assess the correctness of a transition. For instance, if a transition such as ``sliced $\rightarrow$ whole" is not present in FOON, it would be identified as incorrect. Functional units that successfully pass the verification criteria are then added to the mini-FOON.

\subsection{Creating a super-FOON}

We integrate the mini-FOON with the original FOON, forming a combined network known as the super-FOON. During this merging process, our primary focus is on node consolidation, as the mini-FOON and FOON may use different names for the same object or motion node. To achieve consolidation, we follow a set of basic rules. For instance, we convert all object names to their singular form. We observed that GPT-3 often generates plural forms such as "strawberries" and "onions," while FOON represents them as "strawberry" and "onion" respectively. By applying these rules, we try to ensure consistency and compatibility between the node names in the mini-FOON and FOON within the super-FOON network. 

\subsection{Task tree retrieval}

\begin{figure*}[!ht]
     \begin{subfigure}[t]{0.49\textwidth}
         \includegraphics[width=\textwidth]{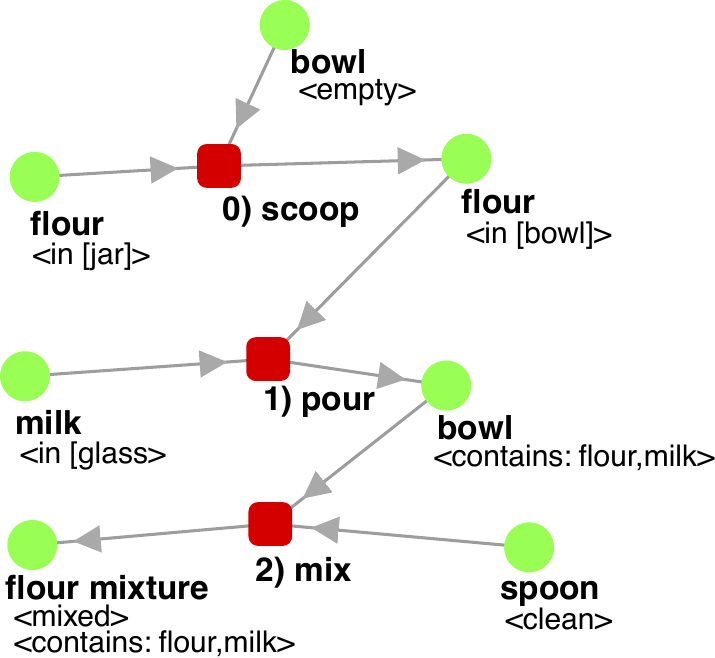}
         \caption{cost of execution = 0.7}
         \label{fig:scoop1}
     \end{subfigure}
     \begin{subfigure}[t]{0.49\textwidth}
         \includegraphics[width=\textwidth]{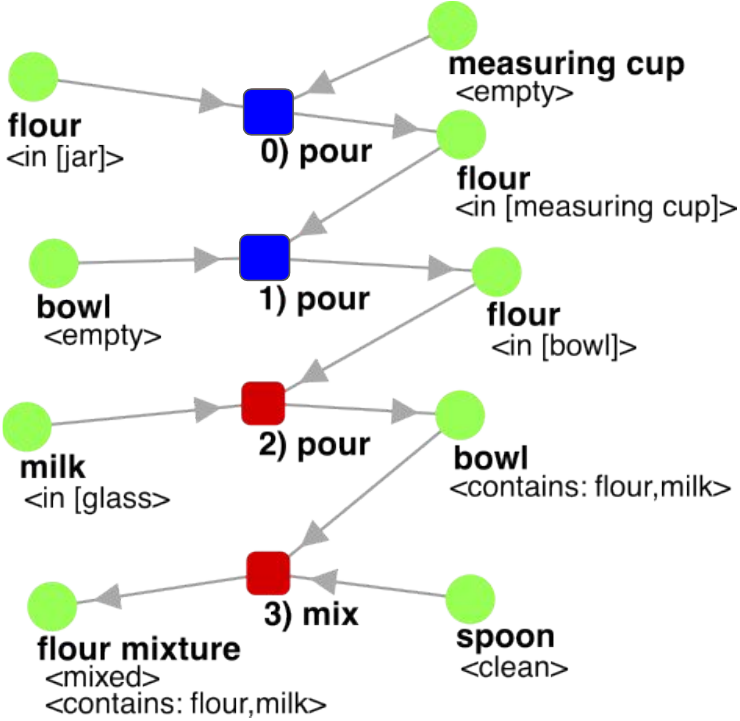}
         \caption{cost of execution = 0.5}
         \label{fig:scoop2}
     \end{subfigure}
     \caption{Example of cost optimization: Comparison between task trees retrieved from the mini-FOON and super-FOON. The assigned costs for scooping, pouring, and mixing are 0.4, 0.1, and 0.1 respectively. (a) The task tree from the mini-FOON (b) The task tree from the super-FOON.}
    \label{fig:example}
\end{figure*}

Taking the desired dish as the goal node, we employ a search procedure similar to \cite{paulius2021weighted} to retrieve all paths leading to the goal. We execute the same search algorithm in both the mini-FOON and super-FOON. This approach often yields multiple task plans, exceeding five in number, which may differ in the number of cooking steps involved. For instance, when preparing a banana milkshake, one plan may suggest adding the whole peeled banana to the blender, while another plan may propose cutting the banana in half before blending.
Once the incorrect functional units have been filtered out, the task tree retrieval procedure does not select them. Instead, it prioritizes the available correct functional units to construct the task plan. For instance, if the functional unit for ``slicing an apple" is found to be incorrect in the first generated tree but correct in the other four task trees, the search procedure will choose the functional unit of slicing an apple from those four task trees.
From the generated plans, we must select the most feasible one for the robot to execute. The feasibility of executing an action depends on the robot's configuration. For example, a robot with only one hand may find pouring easier than chopping. Consequently, the success rate of executing a task tree varies among different robots. Following the approach of \cite{paulius2021weighted}, we assign a cost value ranging from 0 to 1 to each action. These values are determined by three factors: 1) the physical capabilities of the robot, 2) its past experiences and ability to perform actions, and 3) the tools or objects that the robot needs to manipulate. 
A higher cost value indicates a more challenging action to execute. 
Based on these costs, we select task tree 6 from the mini-FOON and task tree 7 from the super-FOON. Ideally, task tree 7 should never be worse than task tree 6 since the super-FOON encompasses all the task trees from the mini-FOON. Task tree 7 serves as the final output of this pipeline. Figure \ref{fig:example} illustrates an example of cost optimization using the super-FOON, where two pouring actions are preferred over scooping due to the significantly lower cost assigned to pouring (0.1) compared to scooping (0.4). We assigned a low cost to pouring based on the successful pouring accuracy achieved by Huang et al. \cite{HUANG2021103692} with a robot.

\section{Experiments and Results}

Our experiment aims to assess both the quality of the generated task trees and the associated execution costs. Simultaneously, we seek to compare the performance of our model in generating recipes across different dish categories. To accomplish this, we curated a dataset consisting of 60 randomly selected recipes from the Salad, Drink, and Muffin categories. These recipes were extracted from Recipe1M+ \cite{marin2019learning}, a comprehensive collection of over one million recipes encompassing a wide range of dish types and ingredients. 

\subsection{Evaluation Metric}

\begin{figure}[!ht]
	\centering
        \includegraphics[width=\columnwidth]{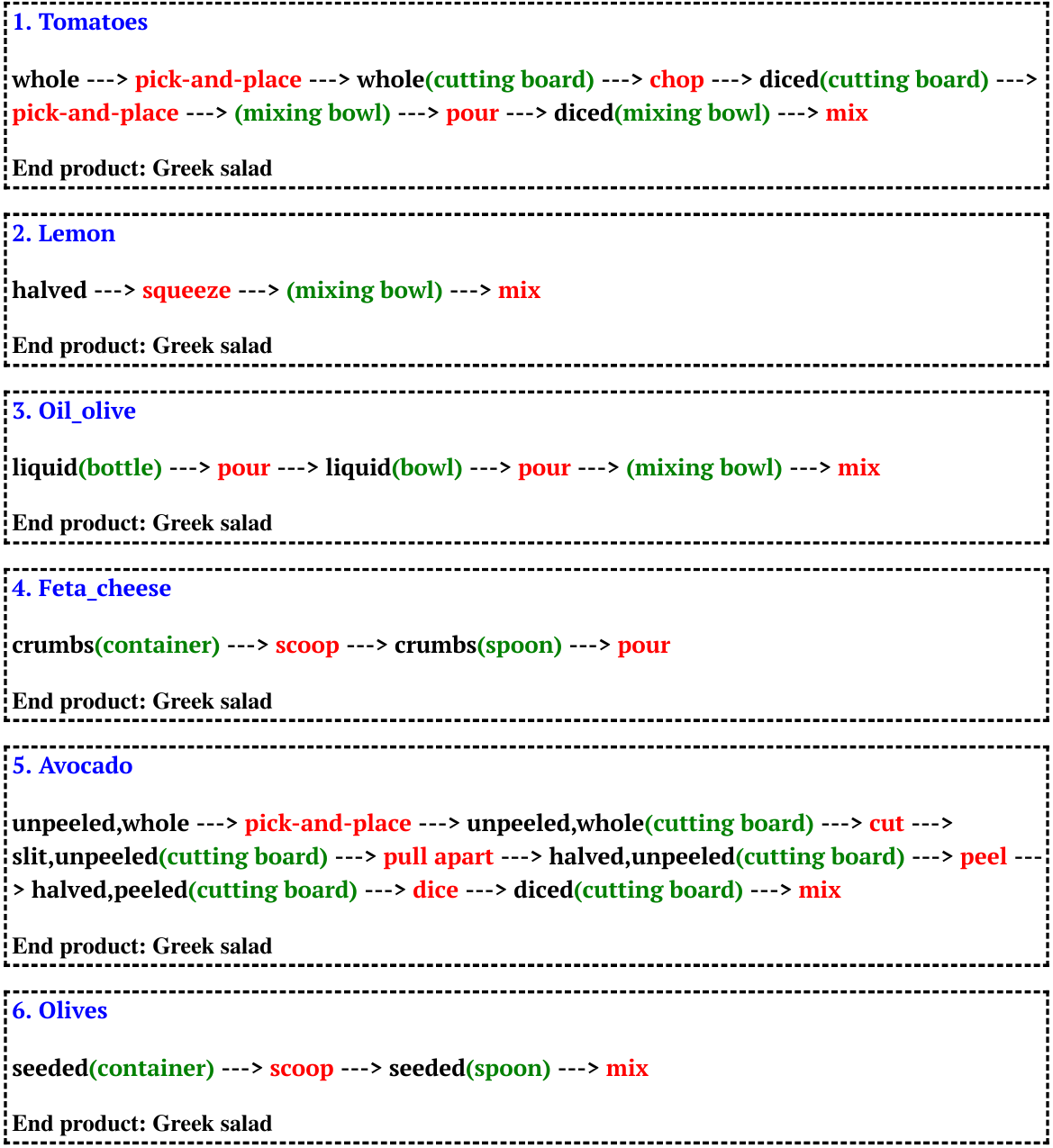}
	\caption{
        Progress lines for a Greek Salad recipe. 
	}
	\label{fig:progress_line}
        \vspace{-0.1in}
\end{figure}

Validating the plan of a cooking task in an automated manner is challenging due to the absence of a fixed method for preparing a dish. Two task plans for the same dish can differ in their cooking approaches, yet both can be deemed correct. As a result, manual verification becomes necessary. However, the original format of a task tree can be difficult for humans to comprehend. To address this, we convert the task trees into progress lines as used in \cite{Sakib2022ApproximateTT} to illustrate how the ingredients are manipulated and undergo changes throughout the cooking process. This simplified visualization facilitates the detection of errors in the task plan by humans. 
We consider a recipe correct if the progress lines for all ingredients used in the recipe are accurate.
An example of progress lines for a Greek Salad recipe is provided in Figure \ref{fig:progress_line}. 

\subsection{Task Planning Accuracy}

\begin{figure}[!ht]
	\centering
        \includegraphics[width=\columnwidth]{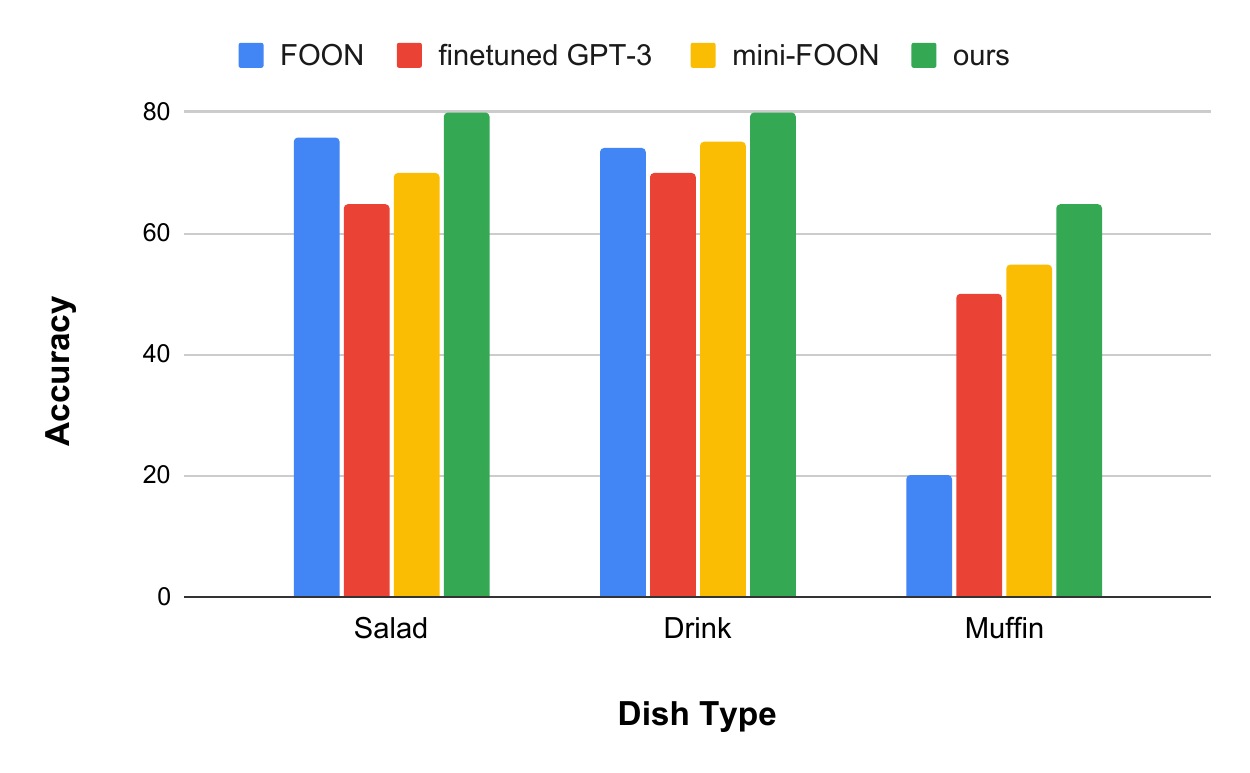}
	\caption{
        Comparison of different approaches’ accuracy on Salad, Drink, and Muffin dishes. 
	}
	\label{fig:comparison}
        \vspace{-.2in}
\end{figure}

We employed four different methods to generate task trees for the selected recipes. The quality of the generated trees was assessed using the progress line, and the corresponding accuracy results are shown in Figure \ref{fig:comparison}.
When relying solely on FOON, the task trees obtained for Salad and Drink recipes exhibited good quality. This was expected as FOON contained an ample number of recipes (10 each) for these categories. However, for Muffin recipes, the quality of the generated task trees suffered due to the scarcity of available examples in FOON (only one recipe). The FOON-search based approach heavily depends on finding a similar recipe in FOON as a reference for making necessary modifications to the task plan. Consequently, a high number of adjustments were required, leading to inaccuracies in the task plan.
In the case of the fine-tuned GPT-3 model, errors in functional units frequently resulted in task plan failures. However, the introduction of the Mini-FOON helped mitigate these errors by providing a wider range of alternatives to achieve the desired objectives. Integrating FOON into our approach enabled us to choose a path from a broader set of options, resulting in higher accuracy.
Compared to \cite{Sakib2022ApproximateTT}, our approach achieved a 4\% higher accuracy for Salad, 6\% higher accuracy for Drink, and a significant 45\% higher accuracy for Muffin recipes. Notably, our fine-tuned model demonstrated good accuracy for Muffin recipes, despite not being specifically trained on this particular dish. This highlights the significant advantage of employing an LLM. Once the LLM is fine-tuned to comprehend the structure of a task tree, it can effectively generalize to various types of recipes.

\subsection{Execution cost}

\begin{figure}[!ht]
	\centering
        \includegraphics[width=\columnwidth]{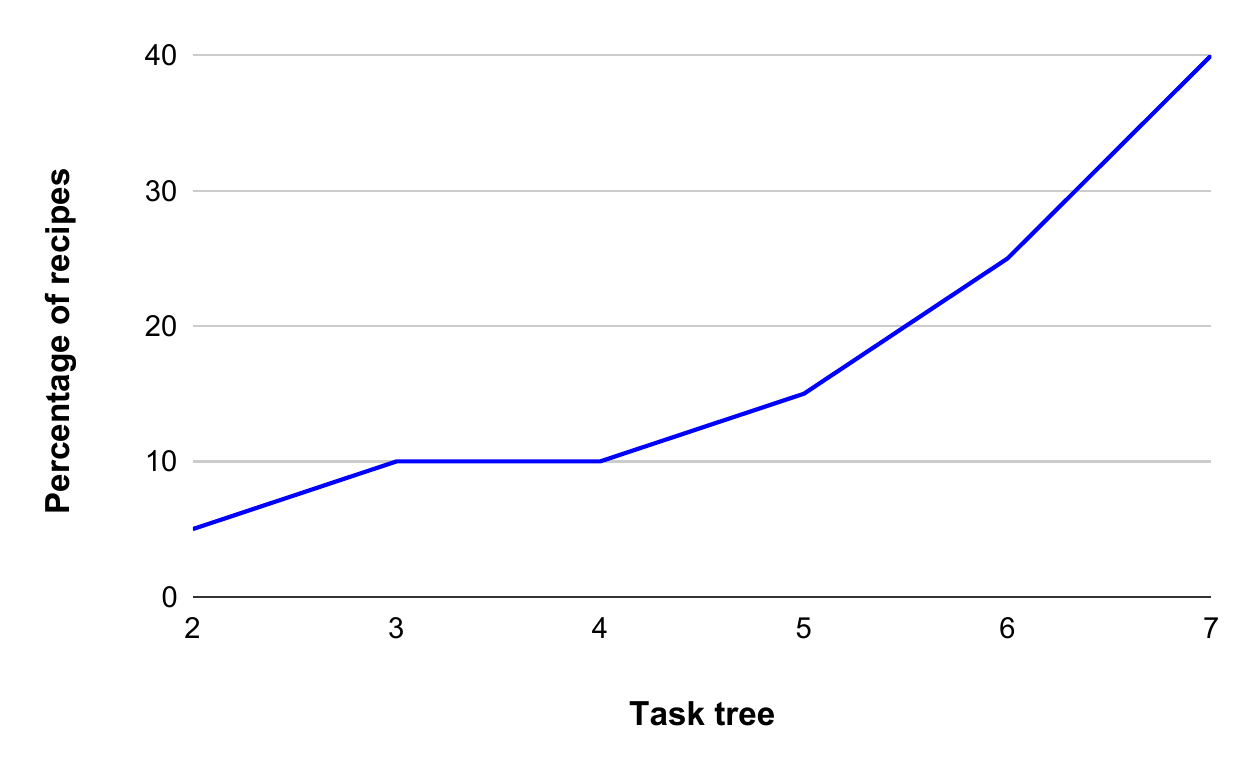}
	\caption{
        Number of recipes that were optimized by generating varying numbers of task trees in comparison to Task Tree 1 (generated by the fine-tuned model).
	}
	\label{fig:cost}
        \vspace{-0.2in}
\end{figure}

The objective of this experiment is to evaluate the extent to which our approach can optimize the execution cost of recipes. If a recipe cannot be optimized, it implies that there are no superior alternatives in FOON compared to the initial output generated by the fine-tuned model (task tree 1). In Figure \ref{fig:cost}, we present the number of optimized recipes by generating different numbers of task trees. When the number of task trees is 2, and we select the plan with the lower cost, it yields a better solution in 5\% of the cases. Similarly, by gradually increasing the number of task trees up to 5 and selecting the one with the minimum cost, we obtain a better solution in 15\% of the cases. More optimization occurs when we choose task tree 6 from the Mini-FOON, as it combines subtasks from five different task trees, resulting in a lower cost. Ultimately, task tree 7, the final output from our pipeline, maximizes the advantages of FOON and minimizes the execution cost compared to task tree 1 in 40\% cases. 


  

\section{Discussion}

\subsection{Finetuning a GPT-3}

\begin{figure}[!ht]
	\centering
        \includegraphics[width=\columnwidth]{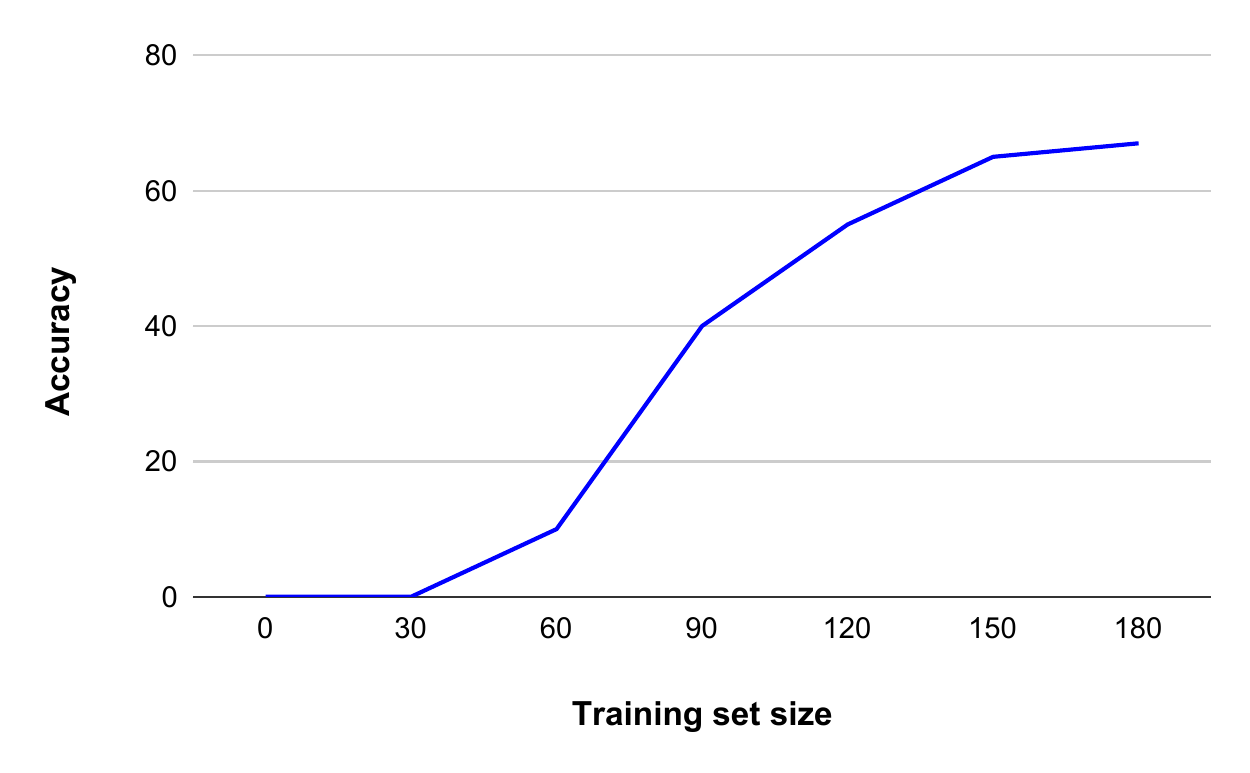}
	\caption{
        Impact of training dataset size on model accuracy.
	}
	\label{fig:finetuning}
        \vspace{-0.2in}
\end{figure}

We examined how the model's understanding of the task tree structure improves with the addition of new training data (Figure \ref{fig:finetuning}).
Initially, the training began with a dataset consisting of only 30 examples. Consequently, the model struggled to grasp the syntax of functional units, resulting in grammatical errors in the generated functional units. For instance, it would include multiple motion nodes within a single functional unit, whereas, according to the definition, a functional unit should contain only one motion node.
As we increased the number of recipes, the model gradually reduced its syntactical errors. However, it still exhibited logical mistakes, such as incorrect state transitions or missing actions.
Finally, after finetuning with 180 examples, the model achieved an accuracy of 67\%. 

\subsection{Executing a task tree}

A task tree provides a high-level plan that lacks interaction with the environment. However, executing actions often requires additional information, such as the geometric position of objects or the initial quantity of ingredients in a container. For instance, a task plan might involve adding ice to an empty glass, but the glass could be positioned upside down on a table. Therefore, before pouring the ice, the glass would need to be rotated back to its original position. This crucial step is missing in our high-level planning. Hence, there is a need for hierarchical planning, where the task tree can be converted into a low-level plan that can be executed in the real world.
Paulius et al. \cite{Paulius2022LongHorizonPA} proposed a method to convert a task tree into a representation using Planning Domain Definition Language (PDDL) \cite{McDermott1998PDDLthePD}. Each functional unit is treated as a planning operator, and a plan is generated based on the robot's low-level motion primitives. 

\subsection{Limitations of our approach}

(i) The generation of a task tree involves making 5 API calls to the fine-tuned model. Each API call takes approximately 5 seconds, resulting in a slow pipeline. The focus of this research was not on time complexity. In the future, if we aim to enhance the system's speed, it may be necessary to explore fine-tuning locally installed LLMs.
(ii) The generated plan sometimes introduces new names for ingredients, states or motions such as garnish. These unknown labels in functional units pose a challenge when attempting to find alternative options in FOON, as proper mapping to existing functional units becomes difficult. Furthermore, the detection of incorrect transitions is also hindered, as the possible transition list may not include these new labels.
(iii) A fine-tuned GPT-3 model has a limitation where the combined query and completion cannot exceed 2048 tokens. Due to this constraint, generating a task tree becomes challenging when dealing with complex recipes that require a higher number of functional units.

\section{Conclusion}

In this study, our objective was to propose a novel pipeline for task tree generation, leveraging the advantages offered by LLMs. We utilized ChatGPT to respond to user queries, and then fine-tuned a GPT-3 model to convert the response into a task tree representation. To enhance the accuracy and execution cost of the task tree, we integrated the output of the fine-tuned model with FOON, exploring multiple possibilities to achieve the desired objectives. Through our experiments, we demonstrated its superior performance, highlighting its remarkable generalization capabilities.
In future, we intend to focus on addressing the challenges of task tree correction and re-planning in cases of planning or execution failures. It is worth noting that our pipeline exhibits a high degree of flexibility, allowing for the seamless substitution of GPT and FOON with more advanced Language Models or knowledge networks. We aim to incorporate image inputs into our system by utilizing the newly released GPT-4, which can handle both textual questions and accompanying images. This would allow users to upload images of dishes and inquire about their preparation methods.


\bibliographystyle{unsrt}
\bibliography{ref}

\end{document}